\documentclass[10pt,twocolumn,letterpaper]{article}

\usepackage{comment}
\usepackage[pagenumbers]{cvpr} 

\usepackage{lipsum}
\usepackage{svg}
\usepackage{float}
\usepackage{placeins}
\usepackage{tabularx}
\usepackage{booktabs}

\definecolor{cvprblue}{rgb}{0.21,0.49,0.74}
\usepackage[pagebackref,breaklinks,colorlinks,allcolors=cvprblue]{hyperref}

\usepackage{soul}

\newcommand{\bpara}[1]{\vspace{0.25em}\noindent\textbf{#1}}

\def\paperID{3441} 
\def\confName{CVPR}
\def\confYear{2025}

\title{Bias for Action: Video Implicit Neural Representations with Bias Modulation}

\author{Alper Kayabasi$^{1}$,\; Anil Kumar Vadathya$^{3}$,\;Guha Balakrishnan$^{2}$,\; Vishwanath Saragadam$^{1}$ \\
$^{1}$University of California Riverside, $^{2}$Rice University \\
$^{3}$ Neal Cancer Center, Houston Methodist Hospital \\
{\tt\small \{akaya003,vishwanath.saragadam\}@ucr.edu,\; guha@rice.edu,\;avadathya@houstonmethodist.org}
}

\begin{document}
\maketitle
\begin{abstract}
We propose a new continuous video modeling framework based on implicit neural representations (INRs) called \textbf{ActINR}. At the core of our approach is the observation that INRs can be considered as a learnable dictionary, with the shapes of the basis functions governed by the weights of the INR, and their locations governed by the biases. Given compact non-linear activation functions, we hypothesize that an INR's biases are suitable to capture motion across images, and facilitate compact representations for video sequences. Using these observations, we design ActINR to share INR weights across frames of a video sequence, while using unique biases for each frame. We further model the biases as the output of a separate INR conditioned on time index to promote smoothness. By training the video INR and this bias INR together, we demonstrate unique capabilities, including $10\times$ video slow motion, $4\times$ spatial super resolution along with $2\times$ slow motion, denoising, and video inpainting. ActINR performs remarkably well across numerous video processing tasks (often achieving more than 6dB improvement), setting a new standard for continuous modeling of videos.
\end{abstract}

\section{Introduction}
\label{sec:intro}
An Implicit Neural Representations (INR) aims to approximate a continuous signal by fitting a multi-layer perceptron (MLP) with non-linear activation functions to the signal. INRs compactly encode the signal into the weights of an MLP, establishing a mapping from input coordinates to corresponding output values. For instance, a video INR may map a spatiotemporal $(x,y,t)$ coordinate to an output RGB color value. By preserving the main characteristics of signals in a smooth manner, INRs serve as versatile tools capable of performing interpolation~\cite{ff-nerv,boosting_nerv}, inpainting~\cite{ds-nerv,d-nerv}, and denoising~\cite{nerv,wire}. Owing to these noteworthy features, INRs have been successfully applied in diverse applications, including novel view synthesis~\cite{nerf,novel-view2}, inverse problems~\cite{inverse}, image generation~\cite{image_gen1}, physics simulation~\cite{physic}, and shape representation~\cite{shape1,shape2}. 

\begin{figure}[!tt]
\centering
\includegraphics[width=\linewidth]{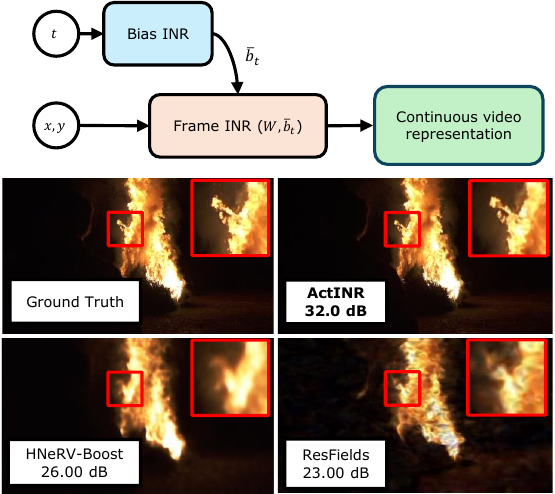}
\caption{\textbf{ActINR: Continuous Video Representation via Bias Modulation.} We propose a new continuous representation of videos called ActINR. At its core is a frame INR that takes continuous spatial coordinates, and a bias INR that takes continuous time coordinates, and outputs bias values for the frame INR at that instant. This bias modulated framework enables us to represent videos in a concise manner, enabling several video processing applications. Shown above is a $10\times$ video slow-motion generated through temporal interpolation. ActINR achieves considerably higher PSNR, with visually pleasing interpolated results.}
\label{fig:teaser}
\end{figure}

\begin{figure*}[!tt]
\centering 
\includegraphics[width=\textwidth]{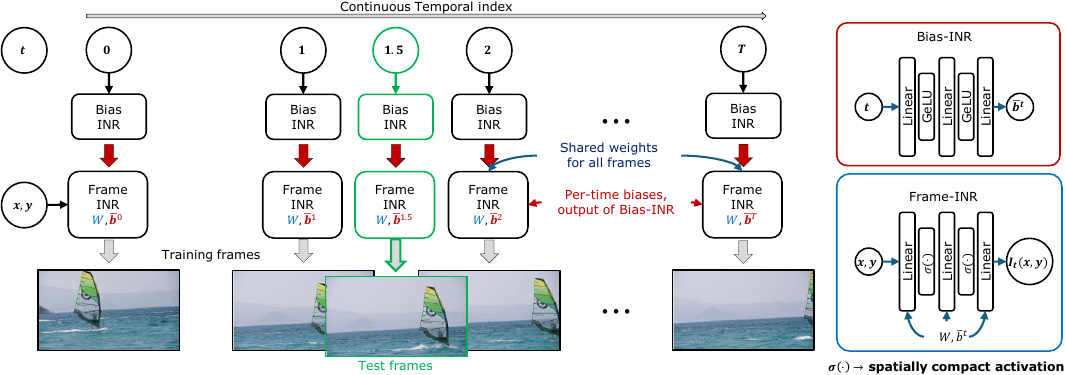}
\caption{\textbf{Schematic overview of ActINR framework}. 
ActINR is a novel video implicit representation model that consists of a frame INR with shared weights across all frames, that takes continuous coordinates as input, and outputs the intensity at that location, and a bias-INR, that takes a continuous time index as input, and outputs the biases for the frame INR at that time index. ActINR is therefore able to model the video continuously both in space and time, enabling slow-motion video generation, video super resolution, denoising, and inpainting.}
\label{fig:arch}
\end{figure*}

INRs have more recently been explored for representing videos, starting with the seminal work of NeRV~\cite{nerv} and followed by several works that expanded upon its speed and representation capacity~\cite{hnerv,denerv,enerv,nirvana,hinerv,gop-nerv,ds-nerv}. Crucial to modeling videos is the ability to capture motion at various spatiotemporal scales, from minute to large displacements per frame and across time. However, existing video INRs still often struggle at modeling patterns of large spatial or temporal distance, as evidenced by their poor performance on dramatic interpolation tasks. For this reason, works such as D-NeRV~\cite{d-nerv} that leverages cross-frame constraints, and FF-NeRV~\cite{ff-nerv} that leverages inter-frame optical flow, address the challenges associated with motion, but invariably fall short on modeling large motions. 

 In this paper, we propose a ActINR (pronounced Actioner), a new implicit representation model for videos that can model small and large motions alike, can robustly handle noise, and is capable of significant super-resolution both spatially (increasing frame resolution) and temporally (slow-motion generation) (see Fig.~\ref{fig:teaser}). At the core of our contribution is the observation that an INR can be thought of a learned dictionary~\cite{structured_dictionary,mitchell}, where the basis functions are modeled by the specific choice of activation function. In this interpretation, the INR's weights model the shape and size of the basis functions, while its biases control their locations. Assuming compact basis functions, e.g., wavelets~\cite{wire} or Gaussians~\cite{gaussian}, a local motion can be modeled by modifying the location of its corresponding basis function, which translates to modulating the values of the INR's \emph{biases}. 

ActINR represents a video by sharing weights of the INR across all frames, while using unique biases per frame. To enable interpolation between frames, ActINR models the biases as the output of a hypernetwork that takes a continuous time index as input. An overview of ActINR architecture is provided in Fig.~\ref{fig:arch}. Further, similar to the KiloNeRF framework~\cite{kilonerf} and JPEG and H.264 compression algorithms, we enable speed up via block processing, by assigning smaller INRs to patches within each frame.

We evaluate ActINR on several video-based inverse problems. We demonstrate excellent performance for $10\times$ slow-motion video generation (up to 5dB better than competition), and video super resolution with $2\times$ slow motion, and $4\times$ spatial upsampling (6dB better than competition). Further, we demonstrate that ActINR acts as a robust prior for video processing by showcasing denoising in extreme conditions (achieving up to 10dB higher than competition), and inpainting (up to 2dB better) across a wide variety of video sequences. Extensive qualitative and quantitative evaluations place ActINR as a strong state-of-the-art baseline for low-level video processing tasks.

\section{Prior Work}

We provide a brief overview of implicit neural representations (INRs) as they are applied to video representations.

\bpara{Implicit neural representation.}  INRs provide a continuous and compact neural network-based parameterization of signals. At their core, INRs map coordinates to signal values—whether in the form of images \cite{siren,wire,incode}, videos \cite{video,video2}, 3D shape \cite{3d-shape,3d-shape2}, scene \cite{scene,scene2}, or audio \cite{audio,audio2}—empowering a broad spectrum of applications, such as novel view synthesis \cite{novel_view}, interpolation \cite{interpolation}, denoising \cite{wire}, compression \cite{compression}, inpainting \cite{inpainting}, registration \cite{deform_register_inr}, computational physics \cite{physic} and style transfer \cite{style}.

A key component affecting INRs capacity is its non-linear activation function. The popular activation choice, ReLU, is however ill-suited for INRs due to their spectral bias towards lower frequencies~\cite{spectralBias}. Numerous works have addressed this issue, including positional encoding~\cite{nerf}, sinusoidal activation function~\cite{siren}, Gaussian activation function~\cite{gaussian}, wavelet activation function~\cite{wire}, and many more.

INR can be interpreted as modeling a signal with basis expansion functions governed by its activation \cite{structured_dictionary, mitchell}. A direct implication is that an activation function with compact spatial support enables modeling local motion. We show that the weights of the MLP have a direct effect on the shape of the basis functions, while the biases control their location. Our goal then is to modulate biases to model local motion, enabling a continuous representation of videos.

\bpara{Implicit video representations.} An obvious choice for modeling videos with  INR is to have $(x, y, t)$ as input. However, training such an INR is computationally demanding. Phase-INR \cite{phaseinr} eliminates the time coordinate by substituting it with a phase shift embedded within the positional encoding tied to the spatial domain. ResField \cite{resfield} similarly removes the time coordinate by replacing it with residual weight matrices. 
Injecting time through a single phase is overly simplistic, while incorporating it into a highly parameterized weight matrix is excessively complex. By replacing the time coordinate with biases, our approach strikes a balance and offers a middle ground for accurate and efficient temporal dynamics modeling.

To accelerate the encoding time of INR in similar spirit, Neural Video Representation (NeRV) \cite{nerv} employs a convolutional decoder that maps frame indices to their corresponding frames. Hybrid NeRV \cite{hnerv} enhances this approach by replacing frame indices with embeddings extracted from a convolutional encoder.
D-NeRV~\cite{d-nerv} fuses inter-frame embeddings with frame ones, thereby restoring lost dynamics. Boosting NeRV \cite{boosting_nerv} further refines NeRV's performance and compactness by modulating feature maps using affine transformations. DS-NeRV encodes video dynamics and static elements separately through latent codes with distinct sampling rates: frequent samples capture dynamic parts, while sparse samples model static regions. FF-NeRV \cite{ff-nerv} estimates optical flow between the central frame in a group and its surrounding frames. To represent the central frame, it warps the surrounding frames with corresponding flows and aggregates the warped versions with the current frame prediction. Furthermore, while all NeRV-based methods trade off encoding and decoding speed for the flexibility to query pixel values at any spatial position, our approach preserves the ability to zoom not only across time but also within space.

\bpara{Space decomposition methods.}
It is worth noting that motion in videos is often local, a property that is leveraged by video codecs \cite{motion-compensation1,motioncompensation2}. This observation inspires an efficient divide-and-conquer approach similar to KiloNeRF~\cite{kilonerf} to represent each frame through spatial decomposition. Daniel \etal \cite{derf} empirically demonstrated that increasing the capacity of INRs yields diminishing returns, thus validating the divide-and-conquer strategy. In this context, DeRF \cite{derf} employs a learnable Voronoi tessellation to segment space, while Jiayi \etal \cite{super-pixel} allocated distinct INR to super-pixels being formed by clustering based on spatial proximity and signal similarity. 
We employ a simple decomposition consisting of disjoint, equal-sized patches in each frame, which enables modeling local motion more efficiently, and makes the problem computationally tractable.

\section{ActINR}
We now present our technique named ActINR. We first motivate our approach by looking at the internal working of an INR, and then describe our approach.

\bpara{INRs and Basis Representation.}
An Implicit Neural Representation (INR) provides a continuous approximation of a target signal \( I: \mathbf{x} \in \mathbb{R}^D \rightarrow g(\mathbf{x}) \in \mathbb{R}^C \) by utilizing a neural network \( f_{\theta}:  \mathbf{x} \in \mathbb{R}^D \rightarrow f(\mathbf{x}) \in \mathbb{R}^C \) composed of linear layers and nonlinear activation functions. The learnable parameter set, including weight matrices and bias vectors, is collectively represented by \( \theta =\{\mathbf{W}^{1},\mathbf{b}^1,\ldots,\mathbf{W}^{L},\mathbf{b}^L\} \). INR is trained to represent the target signal by minimizing the mean-squared error objective between $\mathbf{g}(\mathbf{x}_s)$ and $\mathbf{f}(\mathbf{x}_s)$, where $\mathbf{x}_s$ denotes sampled input-output pairs. Specifically,
\begin{equation}
\begin{aligned}
\mathbf{y}^{(0)} & = \mathbf{x}  \\
 \mathbf{y}^{(l)} & =  
 \sigma(W^{(l)}\mathbf{y}^{(l-1)} + b^{(l)}) \\
f_\theta(\mathbf{x}) &= W^{(L)}\mathbf{y}^{(L-1)} +  b^{(L)} 
\end{aligned}
\label{eq:inr_defined}
\end{equation}

An INR can be conceptualized as a learnable basis-function expansion of the target signal, with the activation function $\sigma$ dictating the type of basis functions employed (e.g., sinusoidal,
gaussian, wavelet), while the weights and biases, $\mathbf{W}$ and $\mathbf{b}$, modulate the intrinsic properties of each component according to the selected activation function~\cite{structured_dictionary, mitchell}. For compactly supported activation functions, $\mathbf{W}$ and $\mathbf{b}$ control spatial extent and the center of each component, respectively. INRs using such activation functions operate by aggregating such components with compact support, thereby providing an independent representation of local regions within the domain. 

\begin{figure}[!tt]
\centering 
\includegraphics[width=\linewidth]{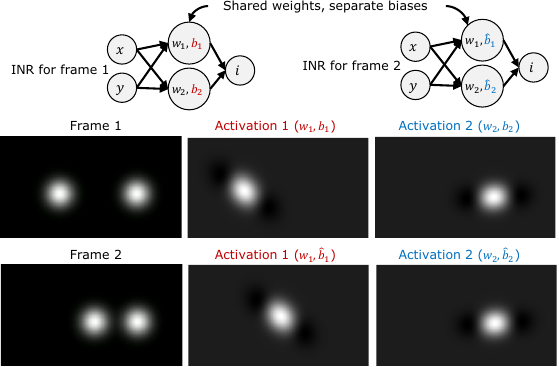}
\caption{\textbf{Toy example motivating the role of biases.} Our toy example illustrates a video featuring two Gaussian blobs: the left blob moves toward the stationary one on the right. Column 1 displays snapshots of the ground truth at two different times, while Columns 2 and 3 show two distinct basis functions in ActINR corresponding to those times. These basis functions evolve over time in response to motion in scene, with their movement governed by biases, thus linking biases directly to motion.}
\label{fig:toy_motion_repr}
\end{figure}

\bpara{Bias-motion Interplay.} Our study is grounded in the hypothesis that \emph{motions in video can be modeled via the movement of compact local basis functions in an INR, governed by the INR's bias parameters.} Furthermore, assuming scene elements are largely the same in appearance over time, the weights of the INR may be shared across time. \cref{fig:toy_motion_repr} illustrates these ideas for a toy video in which a Gaussian blob on the left moves toward a stationary one on the right. While the morphology of the basis functions stays constant over time, basis function \#1 translates rightward in response to the rightward motion of the left-hand Gaussian by changing bias values, while basis function \#2 remains fixed. Basis functions \#1 and \#2 also minimally interfere with each other due to their bounded spatial support, facilitating local motion modeling.

\bpara{ActINR Design.} Based on the intuition above, we now present the proposed ActINR design to model a video sequence. To promote local motion modeling and tractability, we partition an input video into uniformly sized blocks of limited spatial extent and full temporal extent, and fit each block independently. Each layer $l$ in the $L$-layer INR for the $i^{\mathrm{th}}$ patch in the block then has the structure:
\begin{equation}
\begin{aligned}
\mathbf{y}^{(0)}_{i}(x,y) & =[x, y]^T \\
 \mathbf{y}^{(l)}_{i}(x,y) & =  
 \sigma(W^{(l)}\mathbf{y}_{i}^{(l-1)} + b_{i}^{(l)}),
\end{aligned}
\label{eq:normal_inr}
\end{equation}
where $\sigma (\cdot)$ is non-linear activation function, $W^{(l)}$ represents the shared weight matrix across patches, $b_i^{l}$ is the corresponding bias, and $y_i^{l}$ is the output of the layer. The output of the final layer will simply be predicted RGB color values for the input pixel coordinate. We train the INR using a mean-squared error loss over all pixels within the block. In our experiments, we use the wavelet activation function WIRE~\cite{wire} as $\sigma(\cdot)$.

\bpara{Continuous Representation of Biases.} The formulation presented in \cref{eq:normal_inr} assumes independently optimized bias vectors for the input video frames. However, this strategy will likely not result in good temporal interpolation performance at test time. To address this, we predict bias through another INR, thereby enforcing continuity in bias estimates over time. We introduce a bias-INR $\psi$, an MLP with GeLU activation, that outputs bias vectors to be used in each layer of \cref{eq:normal_inr}. More formally, each layer of $\psi$ follows:
\begin{equation}
\begin{aligned}
    \mathbf{y}^{(0)}_t &  =[x,y]^T \\
     \mathbf{y}^{(l)}_t & =  \sigma(W^{(l)}\mathbf{y}^{(l-1)} + \psi^{(l)}(\gamma(t),\mathbf{z}^{l}))
\end{aligned}
\label{eq:bias_inr}
\end{equation}
The formulation has two components: a patch-wise latent code $\mathbf{z}$, and a random fourier feature (RFF) mapping $\gamma(\cdot)$. $\mathbf{z}$ conditions the shared bias-INR across all patches by capturing the degree of immobility within each patch, mitigating the need to assign separate bias-INRs to each patch, reducing the overall parameter count. $\gamma(\mathrm{t})=[\sin{2\pi\mathbf{B}\mathrm{t}},\cos{2\pi\mathbf{B}\mathrm{t}}]^\top$, encodes time as a collection of sinusoidal and cosine signals with frequencies sampled from a Gaussian distribution of variance $\sigma^2$.

\section{Experiments}
We now demonstrate the efficacy of our approach through multiple video-related experiments.

\begin{table*}[!tt]
    \centering
    \begin{tabular}{p{80pt}cccccccc}
    \toprule
    \textbf{DAVIS dataset} & Swan & Camel & Bmx & Bear & Surf & Eleph & Car-turn & Average \\
    \hline 
     ResFields & 21.9/0.56 & 21.3/0.61 & 18.3/0.41 & 22.2/0.64 & 23.9/0.75 & 24.5/0.75 & 23.3/0.66 & 22.2/0.62\\
     FF-NeRV &  21.1/0.56 & 19.4/0.56 & 17.0/0.39 & 20.7/0.60 & 22.2/0.73 & 22.7/0.73 & 22.2/0.66 & 20.8/0.60\\ 
     H-NeRV Boost  &  21.1/0.57 & 20.9/0.66 & 17.3/0.36 & 21.3/0.70 & 22.9/0.76 & 24.7/0.81 &23.2/0.68 &  21.6/0.65\\
     Ours  & \textbf{22.8}/\textbf{0.64}  & \textbf{22.3}/\textbf{0.71} &  \textbf{18.6}/\textbf{0.45} & \textbf{24.0}/\textbf{0.77} & \textbf{23.9}/\textbf{0.78} & \textbf{25.4}/0.81 & \textbf{23.4}/0.68 & \textbf{22.9}/\textbf{0.69}\\
     \bottomrule
    \textbf{UVG dataset} & Bosph & Beauty & Honey & Jockey & Ready & Shake & Yacht & Average \\
    \hline 
    ResFields & 33.2/0.93  & 31.3/0.86  & 35.3/0.97 & 22.7/0.74 & 24.1/0.79 & 29.5/0.86 & 27.6/0.87 & 29.1/0.86 \\
     FF-NeRV  & 35.0/0.96 & 31.0/0.87 & \textbf{39.2}/\textbf{0.98} & 22.2/0.76 & 20.8/0.73 & 29.4/0.87 & 26.5/0.88 & 29.2/0.86\\ 
     H-NeRV Boost  & 36.1/0.97 & 31.5/0.87 & 38.8/0.98 &22.8/0.78
     & \textbf{26.6}/\textbf{0.91} &  30.3/0.88& 28.0/0.91 & 30.6/0.90 \\
     Ours & \textbf{37.3}/\textbf{0.98} & \textbf{31.6}/\textbf{0.87} & 37.6/0.98 & \textbf{24.0}/\textbf{0.78} & 25.9/0.89 & \textbf{30.4}/\textbf{0.88} & \textbf{30.3}/\textbf{0.93} & \textbf{31.0}/0.90 \\
    \bottomrule
     \end{tabular}
     \caption{\textbf{$2\times$ video interpolation on DAVIS and UVG.} The metrics above compare every other frame video interpolation on the DAVIS and UVG datasets. On DAVIS, we observe that our approach achieves on an average of 1dB higher performance in PSNR, and a 0.05 increase in SSIM across the board. For UVG, we observe an average PSNR improvement of 0.4 dB, outperforming all other methods across all scenes except for Honey, which is highly static, and Ready, where we achieve competitive results.}
     \label{tab:davis_interp}
\end{table*}

\setlength\tabcolsep{0pt}
\begin{table*}[!tt]
    \centering
    \begin{tabular*}{\linewidth}{@{\extracolsep{\fill}} cccccccccc}
    \toprule
    \textbf{Video} & Swan & Camel  & Dog  & B-dance & B-trees & C-round & C-shadow & Dance
    & Cows \\
    \hline 
     Ds-NeRV       & 32.3/-          & 27.2/-       &  33.0/-& 31.5/-  & 29.8/- &  29.3/-        & 35.3/-       & 28.8/- & 25.0/- \\ 
     H-NeRV Boost  & 34.1/0.96      & 31.1/0.95        & 33.9/0.96 & \textbf{33.1}/0.98  & 33.0/0.95 &  31.9/0.97        & \textbf{35.8}/0.97 &  30.8/0.92  & 28.3/0.94 \\
     Ours          & \textbf{34.7}/\textbf{0.98} & 31.1/\textbf{0.96} &  \textbf{38.2}/\textbf{0.99} & 32.1/0.98 & \textbf{36.7}/\textbf{0.98} & \textbf{33.3}/\textbf{0.98}  & 35.4/\textbf{0.98}  & \textbf{34.7}/\textbf{0.98} & \textbf{29.9}/\textbf{0.97} \\
     \bottomrule
    \end{tabular*}
    \caption{\textbf{Video inpainting.} The table above compares inpainting results on selected videos from the DAVIS dataset. In each video, we masked five square regions of 50 pixels. Bias-INR achieves an average PSNR improvement of $1.6$dB and an MSSIM increase of $0.03$, along with consistently strong qualitative results. These findings underscore the positive influence of incorporating a motion model as a regularizer in the inpainting.}
    \label{tab:inpainting}
\end{table*}
\setlength\tabcolsep{6pt}

\begin{table*}[!tt]
    \centering
    \begin{tabular}{ccccccccc}
    \toprule 
    \textbf{Video} & Swan  & Dog  & Surf & B-trees & C-round & C-shadow & Dance-t & Average\\
    \hline 
     H-NeRV Boost  & 26.0/0.80      & 24.6/0.74 & 25.3/0.72 & 25.4/0.77 & 26.4/0.85  &  25.2/0.79 &  25.8/0.77  & 25.5/0.78 \\
     Ours      & \textbf{29.1}/\textbf{0.89}   &  \textbf{29.8}/\textbf{0.87}&  \textbf{30.9}/\textbf{0.90} & \textbf{28.7}/\textbf{0.87} & \textbf{27.0}/\textbf{0.88} & \textbf{29.5}/\textbf{0.89}  &  \textbf{28.1}/\textbf{0.87} & \textbf{29.0}/\textbf{0.88} \\
     \bottomrule
    \end{tabular}
    \caption{\textbf{Video denoising.} The table above compares denoising results on some videos from the DAVIS dataset. In each case, we denoised 100 frames by adding readout noise (additive white Gaussian Noise), and shot noise (Poisson), resulting in an input PSNR between $16-18$dB. Across the board, Bias-INR achieved $3-5$dB higher PSNR, with visually pleasing results, emphasizing the appropriateness of our motion model.}
    \label{tab:denoising_quant}
\end{table*}

\bpara{Dataset.} We demonstrate the efficacy of bias-INR across various benchmarks featuring diverse video content, including the UVG dataset \cite{UVG} and the DAVIS dataset \cite{DAVIS}. The UVG dataset comprises seven videos, each with a resolution of 1920×1080 pixels and 300 or 600 frames. The DAVIS dataset consists of 50 videos at 1920×1080 resolution, characterized by substantial motion and intricate spatial details. We evaluate inpainting on a subset of 10 DAVIS videos following the approach in \cite{ds-nerv}. Additionally, we established the first benchmark for denoising using the same DAVIS videos employed for inpainting. For interpolation tasks, we utilized the UVG dataset and selected another subset of seven videos from DAVIS.

\bpara{Evaluation.} We assess the quality of the reconstructed video using two widely accepted metrics, peak signal to noise ratio (PSNR) and MS-SSIM \cite{mssim}, to quantify the level of distortion compared to the original video.

\bpara{Implementation Details.} We employ a three-layer MLP with WIRE activation functions, each hidden layer having a dimension size of 36 {per each spatial block}. {The bias-INR hypernetwork} utilizes Random Fourier Features (RFF) with a size of 40, where frequencies are sampled from a zero-mean Gaussian distribution with a variance of 5. For interpolation tasks, the scale and frequency parameters of WIRE are set to 10 and 100, respectively, while for inpainting tasks, they are set to 5 and 50. The video volume is divided into 96×96 patches, each spanning 10 frames, with a distinct MLP unit assigned to each patch. Each MLP unit is trained for 2000 iterations using the Adam optimizer with a learning rate of 5×$10^{-3}$, coupled with a step decay rate scheduler with a decay ratio of 0.1. Model optimization is carried out by minimizing the mean squared error (MSE) loss between the ground truth and predicted frames. For interpolation and denoising tasks, the model comprises 3.04 million parameters for the DAVIS dataset and 3.17 million parameters for the UVG dataset. All experiments are conducted on NVIDIA RTX 4090 GPU with 24 GB of memory. For more details, please refer to the supplementary material.

\subsection{Video processing with ActINR}
\begin{figure*}[!tt]
\centering 
\includegraphics[width=\textwidth]{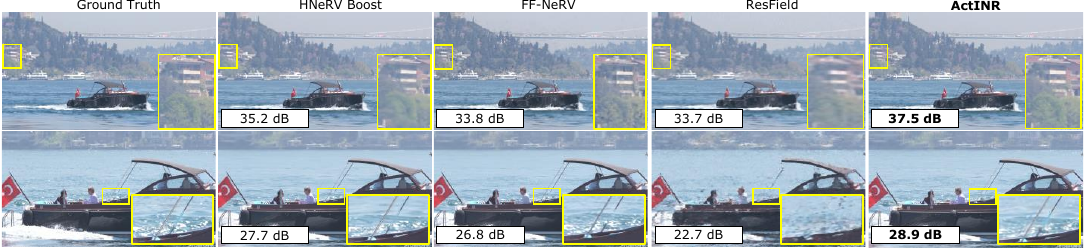}
\caption{\textbf{Qualitative results for $2\times$ interpolation}. In the first row, all methods except ours fail to capture fine-grained details in the video, such as the tree leaves, whereas our approach clearly delineates these intricate elements. In the second row, all other methods struggle to reconstruct the iron rod, blending it with the sea in the background, while our method preserves the rod’s distinct structure and integrity.}
\label{fig:uvg_interp_qual}
\end{figure*}

\bpara{Video Interpolation.}
We evaluate the interpolation capabilities of our method against two state-of-the-art image-wise implicit representation baselines, FF-NeRV and H-NeRV Boost, and one pixel-wise implicit representation, ResField, using the UVG and DAVIS datasets. In \cref{tab:davis_interp}, we exclude every other frame during training, starting from the first, and assess interpolation performance on the unseen frames. 
We observed that image-wise implicit representation methods exhibit inferior performance compared to ours, which we attribute to three primary factors. First, FF-NeRV struggles to accurately estimate optical flow between frames, particularly under conditions of high motion. Secondly, both H-NeRV Boost and FF-NeRV interpolate features for unseen frames, resulting in ghosting artifacts and a loss of detail (e.g. tree branches in \cref{fig:uvg_interp_qual}).

Lastly, convolutional reconstruction introduces a locality bias, resulting in the smoothing of finer details. For example, both image-wise methods fail to capture the leaves of the tree, depicting it merely as a green blob in the second column of \cref{fig:uvg_interp_qual}, whereas our method effectively portrays the granular details of the tree. We also observe that the pixel-wise representation, ResField, underperforms in the interpolation task compared to all other methods. A potential explanation is the over parametrized weight space, making the optimization challenging.

\begin{figure*}[!tt]
\centering 
\includegraphics[width=\textwidth]{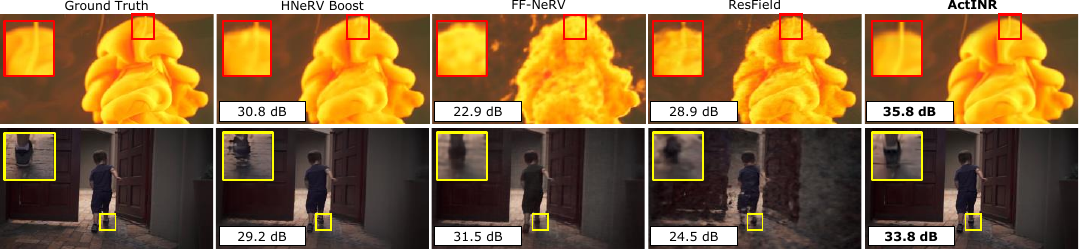}
\caption{\textbf{Qualitative Results for $10\times$ interpolation}. In the first row, our method preserves the integrity of the lava fumes, whereas other methods blend them into the background. In the second row, our approach renders the child’s shoes nearly identical to the ground truth, while other methods suffer from ghosting, loss of detail, and reduced sharpness.}
\label{fig:extreme_interp_qual}
\end{figure*}

\bpara{Extreme interpolation.} To further substantiate our findings, we devised more challenging experimental setups, skipping ten frames instead of every other frame in the training videos, including the ``lava" and ``child" sequences. The qualitative results for this extreme frame-skipping scenario, presented in \cref{fig:extreme_interp_qual}, demonstrate that as the interval between training samples increases, the performance discrepancy between our method and the baselines becomes more pronounced in favor of our approach. We argue that the interaction between the basis functions and the final reconstruction serves as a form of regularization, enabling more precise motion tracking compared to optical flow-based methods, which are prone to errors. While H-NeRV Boost exhibits ghosting artifacts and FF-NeRV fails to resolve motion, resulting in pixel misalignments, our method preserves finer details as illustrated in the final column of \cref{fig:extreme_interp_qual}, despite the extreme frame skipping.

\bpara{Video Inpainting.}
For the inpainting task, we benchmark our method against Boosting NeRV \cite{boosting_nerv} and DS-NeRV \cite{ds-nerv}. We remove five square boxes of size 50 pixels from each video frame. The first box is centered at the center of the frame, while the remaining four are centered within each of the four quadrants specifically at the top-left, top-right, bottom-left, and bottom-right regions. We present quantitative results in \cref{tab:inpainting}, which demonstrate improved inpainting performance over the baselines.  The qualitative results presented in the figure further demonstrate the improvements provided by ActINR. In the figure, while the Boosting Hybrid NeRV reconstructs the cow's spot as extending longer than it should, ActINR delineates the spot better and terminates it exactly where it is supposed to. This is because our method is equipped with not only discontinuity awareness but also motion awareness.

\begin{figure*}[!tt]
\centering
\includegraphics[width=\textwidth]{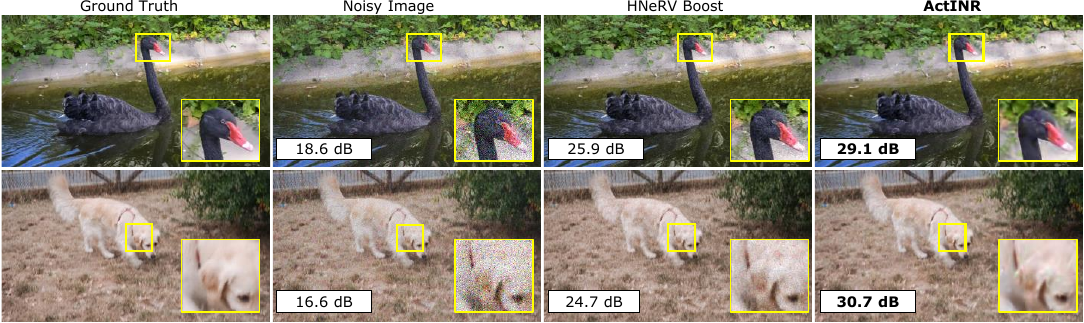}
\caption{\textbf{Video Denoising.} Our model inherently provides implicit regularization, unlike H-NeRV, which directly overfits to video. By constructing its function space with translations of a common building block, it accurately models dynamic evolution. Projecting video into this space effectively rejects noise, as shown in the visual denoising results with at least a $3$ dB improvement in PSNR.}
\label{fig:denoising_qual}
\end{figure*}

\bpara{Video Denoising.} We benchmark our method solely against Boosting NeRV, as it is a state-of-the-art video model that surpasses other methods and has publicly available code. During the denoising process, both our model and Boosting NeRV rely only on supervision from the noisy video. We select photon noise as our noise model due to its realistic representation of sensor behavior. Photon detections adhere to a Poisson distribution with an average rate \( \lambda t \), determined by scene irradiance and integration time. We set the maximum mean photon count of 30, and the readout noise to 5, resulting in an input PSNR of approximately 20 dB. As shown in \cref{fig:denoising_qual}, our method homogeneously filters photon noise and yields a sharp view, whereas HBoostNeRV uniformly retains noise across all regions of the video. We hypothesize that the bias-INR introduces implicit regularization by restricting the biases across frames to evolve smoothly. Quantitative results in \cref{tab:denoising_quant} further demonstrate that our method surpasses the baseline.

\begin{figure}[!tt]
\centering 
\includegraphics[width=\columnwidth]{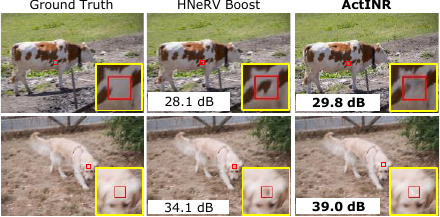}
\caption{\textbf{Video inpainting.} In the first row, HNeRV Boost extends a cow's spot beyond its actual length, and in the second row, it confuses a dog's forehead with its ear. We attribute these error modes to the locality bias of convolution. ActINR, free from locality bias, reconstructs details consistent with the ground truth.}
\label{fig:inpainting_qual}
\end{figure}

\begin{figure}[!tt]
\centering 
\includegraphics[width=\columnwidth]{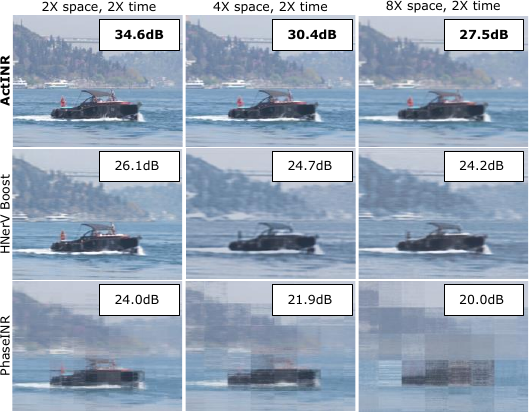}
\caption{\textbf{Spatio-temporal super resolution}. We perform spatial upsampling by factors of 2, 4, and 8, combined with 2x temporal upscaling. Across the board, our method performs significantly better than the baselines, with PSNR gains of 8.5 dB, 5.7 dB, and 3.3 dB for spatial upscaling factors of 2x, 4x, and 8x, respectively.}
\label{fig:space_time}
\end{figure}

\bpara{Space-time super-resolution.} We benchmark our approach against H-NeRV Boost and Phase-INR \cite{phaseinr}. To evaluate super-resolution performance, we conduct experiments on the Bosphorus scene from the UVG dataset across various upscaling factors. Specifically, we set the spatial upscaling factors to 2, 4, and 8, while maintaining the temporal upscaling factor fixed at 2. To the best of our knowledge, we are the first to explicitly demonstrate that NeRV-based methods are incompatible with space-time super-resolution, being confined to temporal super-resolution alone. This empirical finding highlights the versatility of our method over NeRV-based models. Specifically, our approach retains the hallmark feature of neural representations—the ability to query arbitrary coordinates—in contrast to H-NeRV, while maintaining all the advantages that NeRVs offer for internal generalization. Turning to the realm of pixel-wise INRs, we significantly outperform Phase-INR across all upscaling factors. Unlike our method, Phase-INR defines shifts of basis functions exclusively within the positional encoding layer. We attribute the weaker performance of Phase-INR to its lack of a mechanism for modeling shifts of basis functions across all layers. Both quantitative and qualitative results underscore the unique capability of our method to super-resolve along both spatial and temporal dimensions.

\subsection{Ablation Study}
We now explore some of the design choices for our approach, with specific emphasis on activation function, and interpolation module. Additional ablation studies are provided in the supplementary document.

\begin{figure}
    \centering
    \includegraphics[width=\columnwidth]{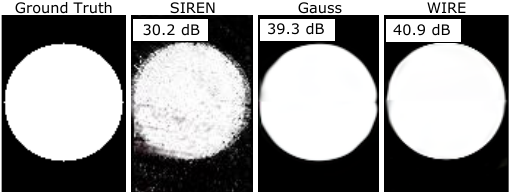}
    \caption{\textbf{Effect of activation function}. The figure shows a toy scene with a circle moving to the right. The result above shows an interpolated frame. While SIREN results in strong spatial artifacts, even in the static background, Gauss and WIRE, with spatially compact activation, exhibit crisp reconstruction.}
    \label{fig:toy_activation_ablation}
\end{figure}

\bpara{Activation choice.}
To evaluate the necessity of compact support, we designed an experimental setup using a synthetic video that depicts a white circle moving consistently to the right against a black background. In this context, we employed INR with three different activation functions: SIREN, Gauss, and WIRE. Our results indicate that SIREN performs the worst as shown in \cref{fig:toy_activation_ablation}, as the network expends its representational capacity to mitigate artifacts resulting from its non-compact support. Visually, the figure reveals anomalies on the black background, where basis functions intended solely for the white circle inadvertently spread background of the image. Although the background does not undergo any motion, it is undesirably influenced by the basis functions modeling the white circle's movement. This phenomenon hinders SIREN's ability to effectively manage local motion, rendering it suitable only for scenarios involving global motion. In contrast, Gauss activation functions achieve the second-best performance. Their lack of an oscillatory structure reduces their representational capacity compared to WIRE, but still allows for better handling of local motion than SIREN. WIRE demonstrates superior performance thanks to its combination of compact and oscillatory characteristics, effectively confining the basis functions to relevant regions and minimizing unwanted influences on the static background.

\begin{table}[t]
    \centering
    \begin{tabular}{ccc}
    \toprule
    \textbf{Interpolation Strategy} & Training & Test \\ 
    \hline
     Oracle       &  46.3 & 46.3 \\
     Linear Interpolation &44.5 & 20.2 \\
     Bias-INR  & 46.0 & 45.8 \\
     \bottomrule
    \end{tabular}
    \caption{The ablation study evaluates an alternative method, specifically linear interpolation, in place of bias-INR. The results justify our choice of INR for continuous bias modeling.}
    \label{tab:interpolation_strategy}
\end{table}

\bpara{Interpolator module.} We devised an experimental framework to validate the essential role of bias-INR in predicting shifts of the basis functions. Specifically, we trained the model on our toy video using \cref{eq:normal_inr}, skipping every other frame, and subsequently interpolated the biases via linear interpolation as an alternative to bias-INR predictions. Furthermore, we trained the model on the toy video by assigning biases to all frames to establish an upper performance bound, referred to as the oracle case. In the linear interpolation scenario, biases are sampled from a learnable lookup table during training, yet linear interpolation itself is not incorporated into the optimization process; it is employed only during the testing phase. This inconsistency between training and testing renders the interpolated values ineffective, as the model cannot adapt the interpolation during inference. Conversely, our approach integrates the interpolator into both the training and testing phases, thereby eliminating any mismatch between the two stages. As shown in \cref{tab:interpolation_strategy}, bias-INR nearly matches the oracle's test performance, while linear interpolation falls significantly short. Specifically, the PSNR drops by 24 dB in the test case compared to training for linear interpolation.

\section{Discussions and Conclusion}
\bpara{Conclusion.} We introduce a novel ActINR to model video data. ActINR shares weights across groups of frames, coupled with a continuous bias representation, allowing the model to effectively and efficiently leverage the inherent temporal redundancy in videos. Extensive experiments reveal that ActINR outperforms state-of-the-art baselines across key downstream tasks, including interpolation, inpainting, and denoising. The inductive biases of the hypernetwork enabled superior reconstruction across all experiments, underlining the efficacy of our approach.

\bpara{Limitation and Future Works.} Our approach assumes that motion remains confined within a designated block size. Consequently, choosing a block that allows objects to exit its boundaries leads to suboptimal performance, and accommodating overflow requires increasing the parameter size proportionally to the magnitude of the motion. Nonetheless, by targeting local and minor motions, we rely on smaller blocks with fewer parameters, which stabilizes the overall parameter count and mitigates the need for larger models. Although our primary focus is on investigating the relationship between biases and motion, we also conducted compression experiments to illustrate performance trade-offs. In these tests, our method performs slightly worse than Hybrid NeRV while outperforming JPEG2000 (see the supplementary for details). Future work will explore sharing parameters across blocks to further address parameter scaling.

\bpara{Acknowledgements} This work was supported by NSF grant CCF-2403123, NIH grant P01IHD109876, the Intelligence Advanced Research Projects Activity (IARPA) via Department of Interior/ Interior Business Center (DOI/IBC) contract number 140D0423C0076. The U.S. Government is authorized to reproduce and distribute reprints for Governmental purposes notwithstanding any copyright annotation thereon. Disclaimer: The views and conclusions contained herein are those of the authors and should not be interpreted as necessarily representing the official policies or endorsements, either expressed or implied, of IARPA, DOI/IBC, or the U.S. Government.
%
\clearpage
\maketitlesupplementary

\begin{figure}[h]
    \centering
    \includegraphics[width=\linewidth]{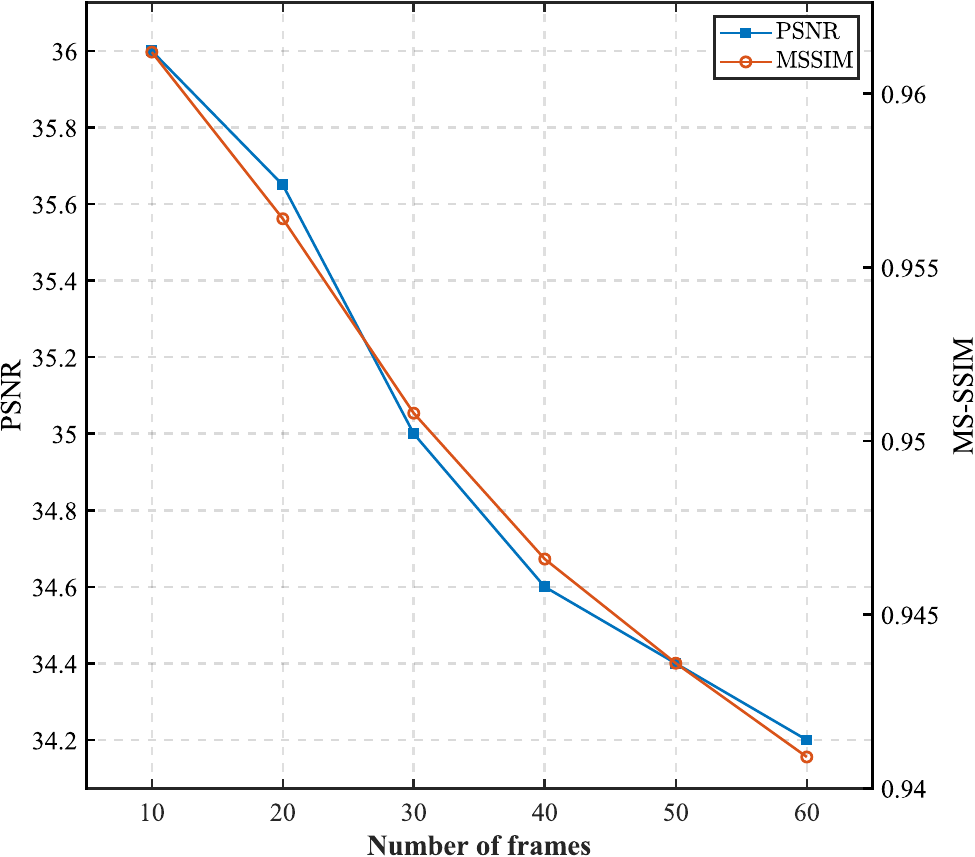}
    \caption{\textbf{Effect of varying GOP size on performance.} The plot reveals a downward trend in reconstruction performance as the number of frames in GOP increases. This finding is consistent with our argument that Act-INR is designed to model local motion.}
    \label{fig:perf_vs_gop}
\end{figure}

\bpara{Performance with varying number of frames.} To validate that Act-INR is designed to model local motion, we conducted experiments using a downscaled version of the Bosphorus video. In these experiments, we increased the number of frames in each group of pictures (GOP) and monitored the performance. As anticipated, the performance deteriorated with a larger number of frames in the GOP as shown in \cref{fig:perf_vs_gop}. This observation aligns with the interpretation that our model maintains strong capacity as long as the object remains within its surrounding box.

\begin{figure}[h]
    \centering
    \includegraphics[width=\linewidth]{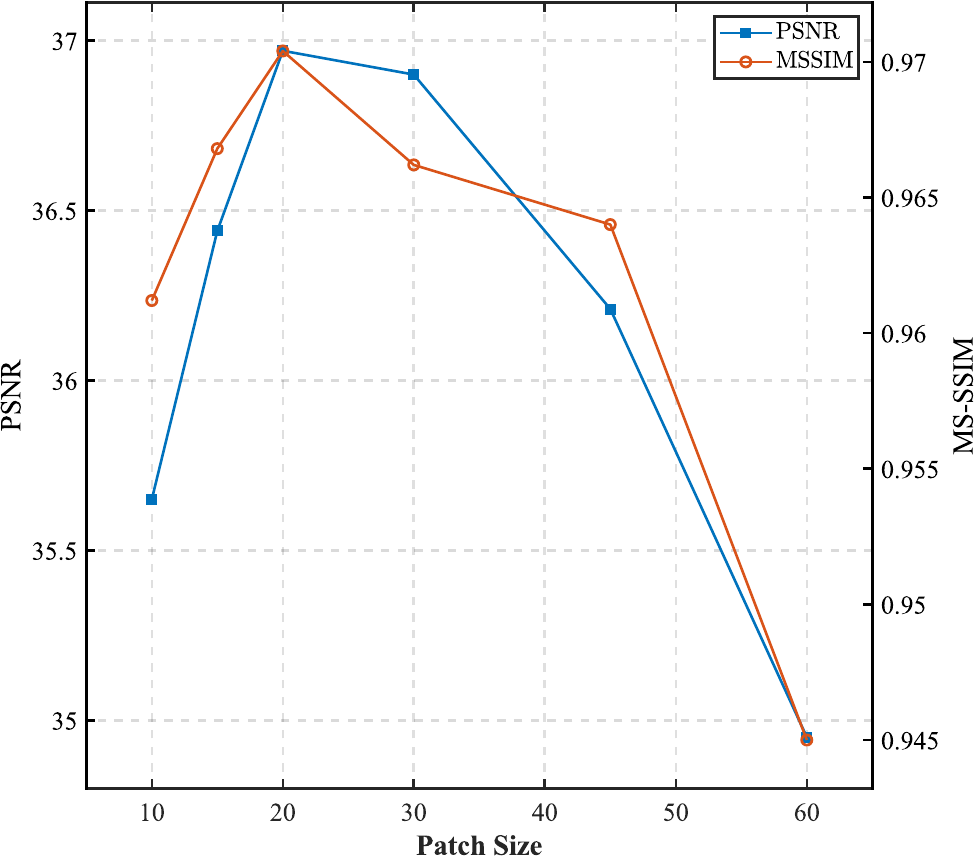}
    \caption{\textbf{Effect of varying patch sizes on performance.} The parameter size is fixed at 1.5 million, and the number of frames in GOP is kept constant at 20. The results demonstrate that a moderate-sized window yields optimal performance.}
    \label{fig:perf_vs_patchsize}
\end{figure}

\bpara{Performance with varying patch size.} In this ablation study, we investigate the effect of patch size on the local motion modeling capability of Act-INR. To ensure a fair comparison, we fix the parameter size and the number of frames in GOP while varying the patch size. This setup allows us to isolate and analyze how changes in patch size influence the model's ability to capture local motion dynamics. Neither excessively large nor excessively small patches allow the model to fully utilize its capacity, as illustrated in \cref{fig:perf_vs_patchsize}. For higher spatial resolution and more complex content, excessively large patches can overwhelm the model, while excessively small patches increase the likelihood of objects moving beyond their designated surrounding box. A moderate patch size serves as an optimal sweet spot, balancing these factors by maintaining manageable spatial resolution and ensuring that objects remain within the designated box for effective modeling

\begin{table*}[h]
    \centering
    \begin{tabular*}{\linewidth}{@{\extracolsep{\fill}} cccccccccc}
    \toprule
    \textbf{Dataset}     &   Bosph & Ready & Yacht & Beauty & Jockey &  Honey  & Shake & Average \\
    \hline
    Ds-NeRV      &  35.2/- & 27.1/- & 29.4/- & \textbf{34.0}/- & 32.9/- & \textbf{39.6}/- &  35.0/- & 33.3/- \\
    H-NeRV Boost &   36.1/0.96  & 30.4/0.91  & 29.3/0.90 & 33.8/0.90 & \textbf{35.8}/0.95 & 39.6/0.98 & \textbf{35.9}/0.96 &  34.4/0.94\\
    Ours        &    \textbf{37.5}/\textbf{0.98}  & \textbf{33.8}/\textbf{0.98}  & \textbf{30.8}/\textbf{0.94} & 33.8/\textbf{0.91} & 34.9/0.95 & 38.4/0.98  & 34.6/0.96 & \textbf{34.8}/\textbf{0.96}\\
    \bottomrule
    \end{tabular*}
    \caption{Video regression results on UVG dataset in PSNR and MS-SSIM}
    \label{tab:video_regression}
\end{table*}

\begin{figure}[h]
    \centering
    \includegraphics[width=\linewidth]{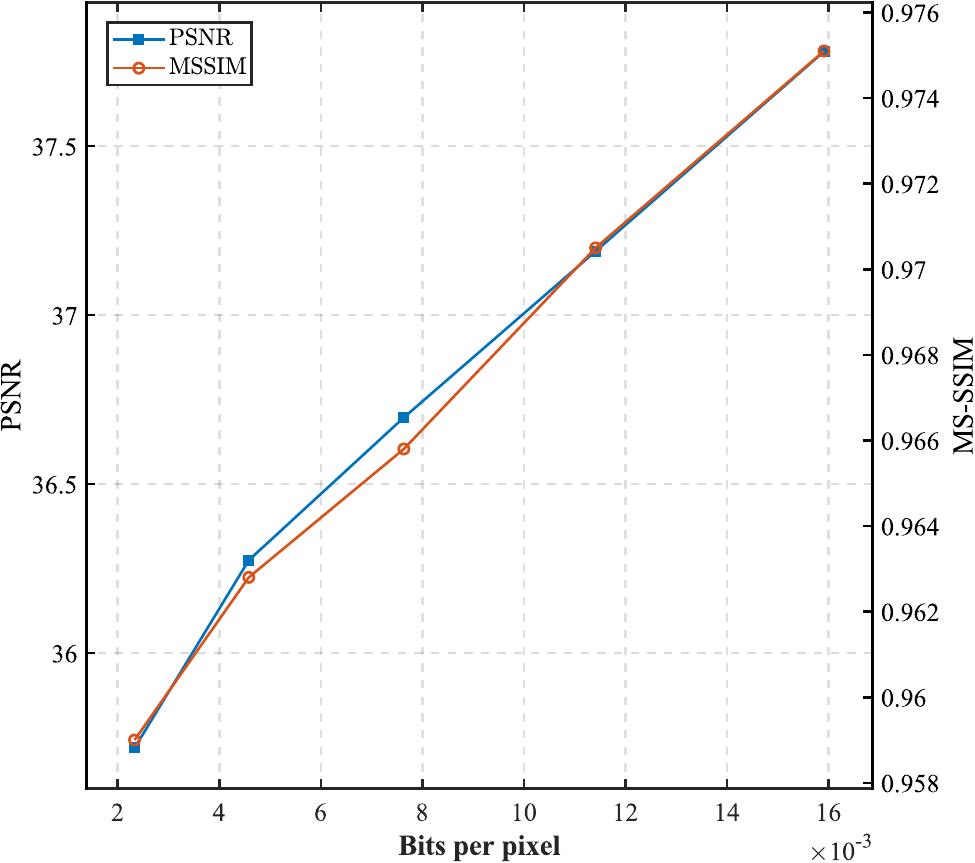}
    \caption{\textbf{Effect of number of parameters on performance.}  The plot shows an increasing trend in reconstruction performance as the parameter count grows. This finding demonstrates that Act-INR scales effectively with parameter size.}
    \label{fig:perf_vs_bpp}
\end{figure}

\bpara{Performance with varying number of parameters.} In this ablation study, we examine how performance scales with the number of parameters. To this end, we progressively increase the feature size from 20 to 60 in increments of 10. The number of frames in GOP is fixed at 20, and the resolution is downscaled by a factor of two, consistent with previous experiments. As shown in \cref{fig:perf_vs_bpp}, the performance of our model improves proportionally with the parameter count, highlighting its scalability.

\bpara{Video regression.}  In \cref{tab:video_regression}, we present a comprehensive evaluation of video regression on the UVG dataset, detailing all PSNR and MS-SSIM metrics obtained. Although the Act-INR architecture is specifically designed for video processing applications, it demonstrates competitive performance compared to state-of-the-art neural representations that are explicitly tailored for video representation tasks.

\begin{figure}[H]
    \centering
    \includegraphics[width=\linewidth]{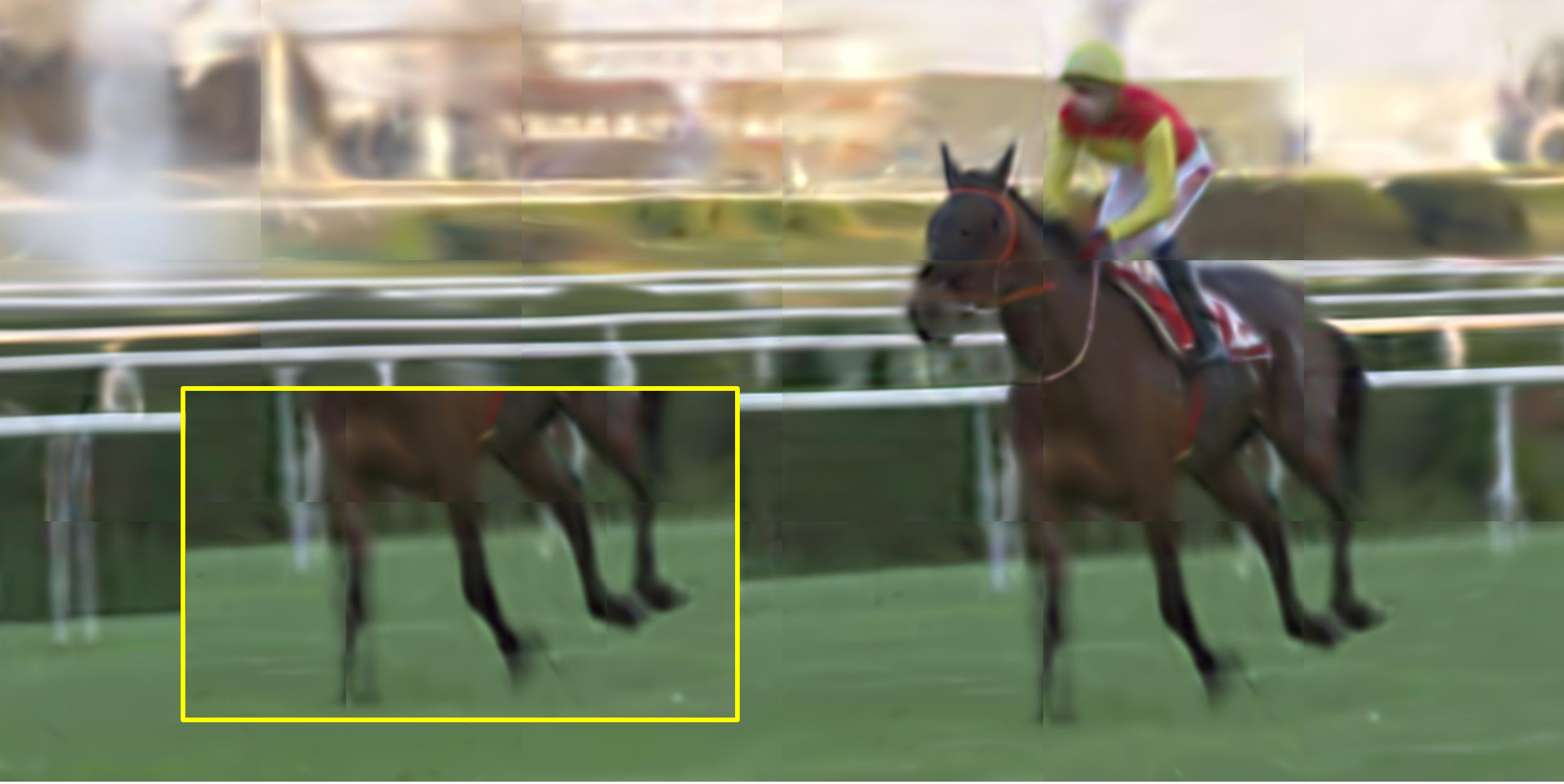}
    \caption{The above illustration highlights a failure case of Act-INR, specifically showing that patch transitions are particularly susceptible to reconstruction artifacts.}
    \label{fig:fail}
\end{figure}

\bpara{Failure case.} In challenging scenarios such as Jockey, we observe that transition regions across patches are not well reconstructed when objects cross patch boundaries. For example, as shown in \cref{fig:fail}, the horse's foreleg remains on the boundary between two patches during its motion. When a patch boundary splits an object into separate parts, the model may struggle to reconstruct these regions in a visually plausible manner.

\begin{figure}[!tt]
    \centering
    \includegraphics[width=\linewidth]{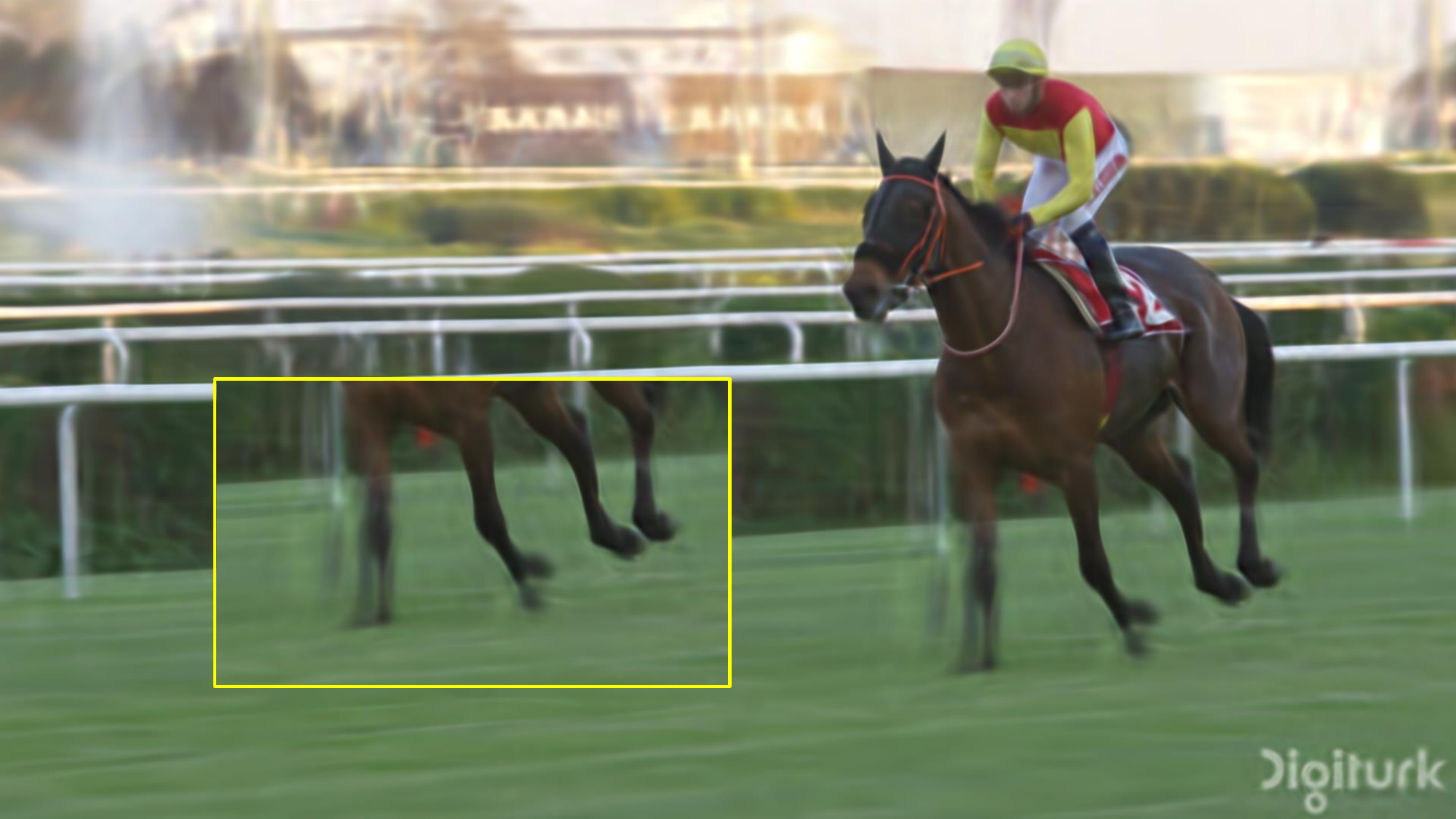}
    \caption{\textbf{Results of training with window blending for the same Video in \cref{fig:fail}.} Overlapping windows effectively eliminate blocking artifacts and improve reconstruction quality at patch transitions.}
    \label{fig:overlap}
\end{figure}

\bpara{Remedy for patch transitions.} To address artifacts arising at patch boundaries when objects cross over patch edges, we propose employing overlapping windows and blending them using a specific strategy, albeit at an increased computational cost. Drawing on the methodology described in Mod-SIREN \cite{window_blending}, pixels are weighted either linearly or bilinearly depending on the number of overlapping patches. Specifically, bilinear blending is applied when a point is surrounded by four windows, whereas linear blending is used when the point is encompassed by two windows. This technique effectively mitigates artifacts at patch boundaries, resulting in smoother reconstructions as shown in \cref{fig:overlap}.

\bpara{Resilience to higher resolution.}
We evaluate the resolution invariance of our interpolation method on four 4K-resolution videos from the UVG dataset. As presented in \cref{tab:high_res}, our method maintains image quality even as the video resolution increases.

\begin{table}[!tt]
    \centering
    \begin{tabular}{ccccc}
     Video   &  Yacht & Ready & Jockey & Bosph  \\
     \hline
     ActINR-4K &  30.1 & 25.7 & 24.0 & 37.2 \\
     ActINR-HD &  30.3 & 25.9 & 24.1 & 37.3 \\
     \bottomrule
    \end{tabular}
    \caption{Interpolation performance on 4K and HD resolution}
    \label{tab:high_res}
\end{table}

\begin{table}[!tt]
    \centering
    \begin{tabular}{ccc}
     Method   &  Encoding Time & Decoding FPS  \\
     \hline
     ActINR &  5h19m & 59.28\\
     FF-NeRV & 6h0m &  84.00\\ 
     HNeRV-Boost & 1h53m & 13.15 \\
     \bottomrule
    \end{tabular}
    \caption{Per-video encoding speed, and decoding performance}
    \label{tab:encod_decod}
\end{table}

\begin{figure}[!tt]
    \centering
    \includegraphics[width=\linewidth]{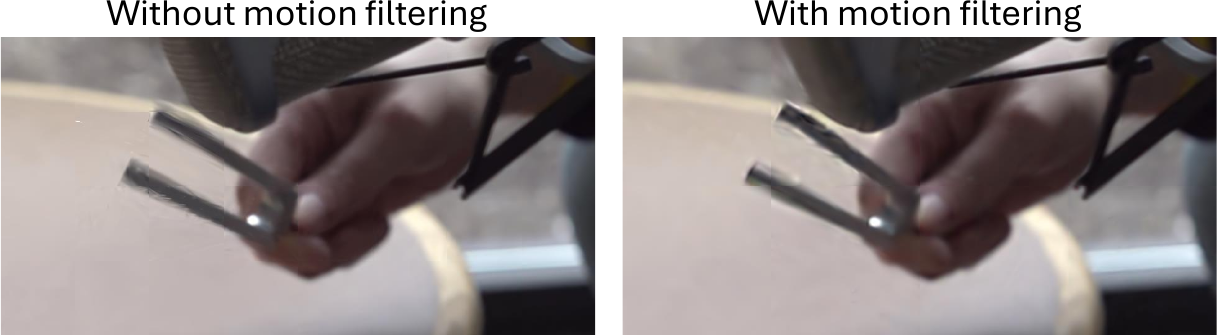}
    \caption{Video of a vibrating tuning fork. Upon applying a median filter to the biases, the pronounced shaking is significantly mitigated, as evident in the image on the right.}
    \label{fig:bias_motion}
\end{figure}

\bpara{Further motivation for bias-motion interplay.}
To emphasize the relationship between biases and motion, we took video of a vibrating tuning fork, and fit our ActINR to it. We then median filtered the time series of biases across time to smoothen the motion artifacts, as shown in Fig.~\ref{fig:bias_motion}. As evident, the high frequency parts around the fork are replaced by a near-static prongs, thereby underscoring the relationship between biases and motions.

\bpara{Encoding and Decoding speeds.} Table.~\ref{tab:encod_decod} tabulates encoding and decoding times for various approaches. Our encoding and decoding times are comparable to previous approaches, with a performance that is similar to FF-NeRV while decoding 4.5× faster than HNeRV-Boost and operating in real time, unlike HNeRV-Boost. We achieve speed up by evaluating all windows in a frame at a time (pytorch \verb|bmm|). We also attribute the speed-up to initializing each group with the preceding group's weights, reducing the training epochs needed for convergence.

\bpara{Further implementation details.} We reproduced FF-NeRV using a 3-million-parameter model for a fair comparison, since the original FF-NeRV reported PSNR only for its 12-million-parameter version. Following the official repository, we kept all hyperparameters unchanged except for model size. Our approach consistently employs a 3-million-parameter model for both interpolation and denoising tasks. However, for inpainting, we used triple the capacity to handle missing regions more effectively.

\begin{figure}[!tt]
    \centering
    \includegraphics[width=\columnwidth]{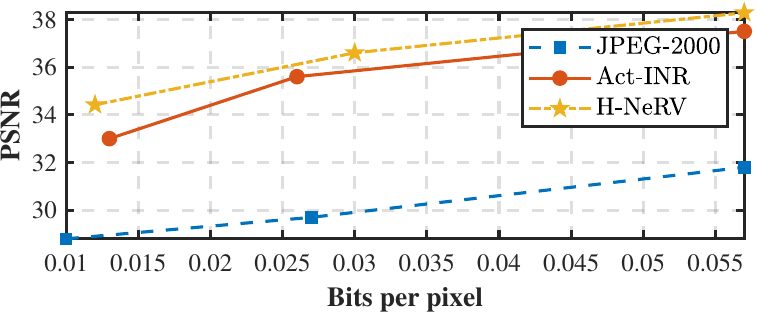}
    \vspace{-1.5em}
    \caption{Rate-Distortion curve for Bosphorus}
    \label{fig:rd}
\end{figure}

\bpara{Compression.} For a fairer comparison, we evaluated the compression performance of Act-INR with simple pruning and 8-bit post-quantization. \cref{fig:rd} shows a rate distortion curve illustrating that Act-INR performs considerably better than JPEG2000, although, slightly worse than Hybrid NeRV. These analyses demonstrate that ActINR primarily excels in solving inverse problems (interpolation, super resolution, denoising, and inpainting) while state-of-the-art techniques like Hybrid NeRV \cite{hnerv} are better suited for compression tasks.
{
    \small
    \bibliographystyle{ieeenat_fullname}
    \bibliography{main}
}


\end{document}